\documentclass{article}

\usepackage{microtype}
\usepackage{graphicx}
\usepackage{booktabs} 
\usepackage{listings} 
\usepackage{multirow} 

\PassOptionsToPackage{hyphens}{url}\usepackage{hyperref}
\usepackage[accepted]{sysml2019}

\usepackage{subfig}

\usepackage{enumitem}

\usepackage{sysml2019}

\graphicspath{ {./figs/} }

\sysmltitlerunning{FixyNN: Efficient Hardware for Mobile Computer Vision via Transfer Learning}

\begin{document}

\twocolumn[
\sysmltitle{FixyNN: Efficient Hardware for Mobile Computer Vision \\ via Transfer Learning}




\begin{sysmlauthorlist}
\sysmlauthor{Paul N. Whatmough}{arm}
\sysmlauthor{Chuteng Zhou}{arm}
\sysmlauthor{Patrick Hansen}{arm}
\sysmlauthor{Shreyas Kolala Venkataramanaiah}{ari}
\sysmlauthor{Jae-sun Seo}{ari}
\sysmlauthor{Matthew Mattina}{arm}
\end{sysmlauthorlist}

\sysmlaffiliation{arm}{Arm ML Research, Boston, MA, U.S.A.}
\sysmlaffiliation{ari}{School of Electrical, Computer and Energy Engineering, Arizona State University, AZ, U.S.A.}

\sysmlcorrespondingauthor{Paul Whatmough}{paul.whatmough@arm.com}

\sysmlkeywords{Machine Learning, SysML}

\vskip 0.3in

\begin{abstract}
The computational demands of computer vision tasks based on state-of-the-art Convolutional Neural Network (CNN) image classification far exceed the energy budgets of mobile devices.
This paper proposes FixyNN, which consists of a fixed-weight feature extractor that generates ubiquitous CNN features, and a conventional programmable CNN accelerator which processes a dataset-specific CNN.
Image classification models for FixyNN are trained end-to-end via transfer learning, with the common feature extractor representing the transfered part, and the programmable part being learnt on the target dataset.
Experimental results demonstrate FixyNN hardware can achieve very high energy efficiencies up to 26.6 TOPS/W ($4.81 \times$ better than iso-area programmable accelerator).
Over a suite of six datasets we trained models via transfer learning with an accuracy loss of $<1\%$ resulting in up to 11.2 TOPS/W -- nearly $2 \times$ more efficient than a conventional programmable CNN accelerator of the same area.
\end{abstract}
]



\printAffiliationsAndNotice{}  

\section{Introduction}
\label{sec:intro}

Real-time computer vision (CV) tasks such as image classification, object detection/tracking and semantic segmentation are key enabling technologies for a diverse range of mobile computing applications, including augmented reality, mixed reality, autonomous drones and automotive advanced driver assistance systems (ADAS).
Over the past few years, convolutional neural network (CNN) approaches have rapidly displaced traditional hand-crafted feature extractors, such as Haar~\cite{violajones} and HOG~\cite{hog}. 
This shift in focus is motivated by a marked increase in accuracy on key CV tasks such as image classification~\cite{vgg}.
However, this highly desirable improvement in accuracy comes at the cost of a vast increase in computation and storage~\cite{energygap}, which must be met by the hardware platform.
Mobile devices exhibit constraints in the energy and silicon area that can be allocated to CV tasks, which limits the adoption of CNNs at high resolution and frame-rate (e.g. 1080p at 30 FPS).
This results in a gap in energy efficiency between the requirements for real-time CV applications and the power constraints of mobile devices.


\begin{figure}[t]
\centering
\includegraphics[width=0.45\textwidth]{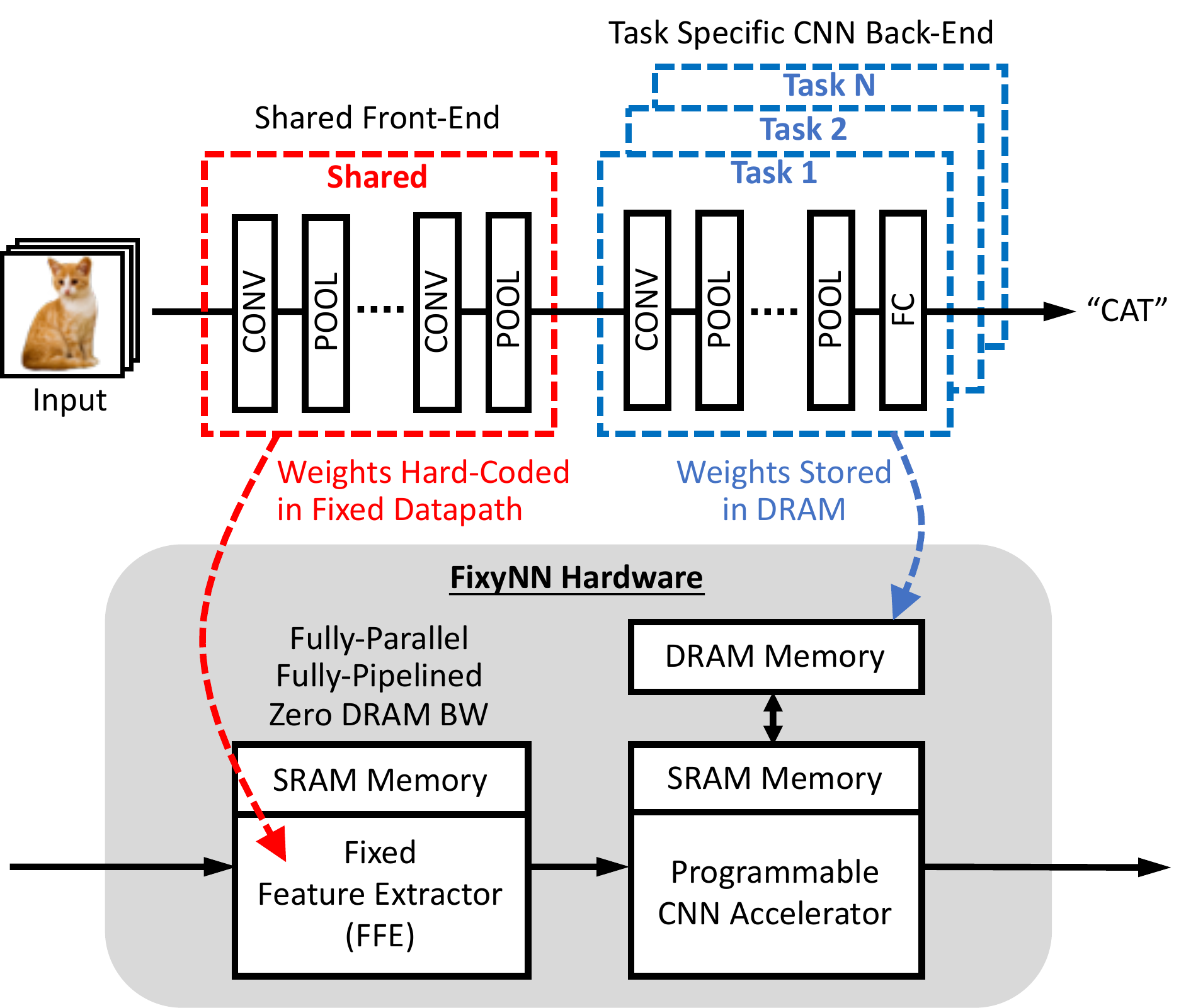}
\vspace{-3pt}
\caption{\textbf{FixyNN} proposes to split a deep CNN into two parts, which are implemented in hardware using a (shared) fixed-weight feature extractor (FFE) hardware accelerator for the shared front-end and a canonical programmable accelerator for the task-specific back-end.}
\label{fig:concept}
\end{figure}

Two key trends that have recently emerged are starting to close this energy efficiency gap: more efficient CNN architectures and more efficient hardware.
The first is the design of more compact CNN architectures.
\textit{MobileNetV1}~\cite{MobileNetV1} was an early and prominent example of this trend, where the CNN topology is designed to minimize both the number of multiply-and-accumulate (MAC) operations and the number of parameters, which is essentially the compute and storage required of the hardware platform.
MobileNetV1 similar accuracy to VGG (top-5 ImageNet 89.9\% vs. 92.7\%), 
with only $\sim$3\% of the total parameters and MACs.
The second trend is the emergence of specialized hardware accelerators tailored specifically to CNN workloads.
Typical optimizations applied to CPU, GPU and accelerators include: provision for small floating-point and fixed-point data types, use of optimized statically-scheduled scratchpad memories (as opposed to cache memories), and an emphasis on wide dot-product and matrix multiplication datapaths. 


In this paper we describe \textbf{FixyNN}, which builds upon both of these trends, by means of a hardware/CNN co-design approach to CNN inference for CV on mobile devices.
Our approach (Figure~\ref{fig:concept}) divides a CNN into two parts.
The first part of the network implements a set of layers that are common for all CV tasks, essentially producing a set of universal low-level CNN features that are shared for multiple different tasks or datasets.
The second part of the network provides a task-specific CNN back-end.
These two CNN parts are then processed on different customized hardware.
The front-end layers are implemented as a heavily optimized \textit{fixed-weight feature extractor (FFE)} hardware accelerator.
The second part of the network is unique for each dataset, and hence needs to be implemented on a canonical programmable CNN hardware accelerator~\cite{nvdla,armml}.
Following this system architecture, FixyNN diverts a significant portion of the computational load from the CNN accelerator to the highly-efficient FFE, enabling much greater performance and energy efficiency.
The use of highly aggressive hardware specialization in the FFE makes FixyNN a significant step forward towards closing the energy efficiency gap on mobile devices.
At the same time, by leveraging transfer learning concepts, we are able to exploit aggressively optimized specialized hardware without sacrificing generalization.

This paper describes and evaluates \textbf{FixyNN}; 
the main contributions are listed below:
\begin{itemize}
\vspace{-10pt}
\item A description of a hardware accelerator architecture for the fixed-weight feature extractor (FFE), including a survey of the potential optimizations.
\item An open-source tool-flow \cite{deepfreeze} for automatically generating and optimizing an FFE hardware accelerator from a TensorFlow description.
\item Demonstration of the use of \textit{transfer learning} to generalize a single common FFE to train a number of different back-end models for different datasets.
\item Present results that compare \textbf{FixyNN} against a conventional baseline at iso-area.
\end{itemize}

The remainder of the paper is organized as follows.
A brief survey of related work is given in Section~\ref{sec:related}.
Section~\ref{sec:hw} highlights the performance and power efficiency advantage of fixed-weight hardware datapaths, and describes our approach to buffering data in fixed-weight layers and our tool flow for automatically generated hardware.
Section~\ref{sec:ml} describes how a fixed feature-extractor can be used with transfer learning principles to train networks for a variety of CV datasets of varying sizes.
Section~\ref{sec:methodology} outlines our experimental methodology, and Section~\ref{sec:results} provides results that combine the hardware and machine learning experiments to show state-of-the-art performance for benchmark tasks.
Section~\ref{sec:conclusion} concludes the paper.

\section{Related Work}
\label{sec:related}

\textbf{CNN Hardware Accelerators.}
There is currently huge research interest in the design of high-performance and energy-efficient neural network hardware accelerators, both in academia and industry~\cite{myriad2,armml,nvdla,reagen_book17}.
Some of the key topics that have been studied to date include 
dataflows~\cite{eyeriss,samajdar_arxiv18}, 
optimized data precision~\cite{minerva}, 
systolic arrays~\cite{tpu}, 
sparse data compression and compute~\cite{eie,cnvlutin,scnn,scalpel,circnn,whatmough_jssc18}, 
bit-serial arithmetic~\cite{stripes}, 
and analog/mixed-signal hardware~\cite{chen_16,redeye,isaac,prime,neurocube,pipelayer}.
There is also published work on hardware accelerators optimized for image classification for real-time CV~\cite{buckler_isca18,riera_isca18,euphrates}, along with simulation tools~\cite{scalesim}.


\textbf{Image Processing Hardware Accelerators.}
The hardware design of the fixed feature extractor in FixyNN is reminiscent of image signal processing hardware accelerators.
In particular, the use of native convolution and line-buffering have been explored in prior works including~\cite{halide,rigel,darkroom,lee_tcomp87,horst_97}.

\begin{figure*}[ht]
\centering
\vspace{6pt}
\includegraphics[width=\textwidth]{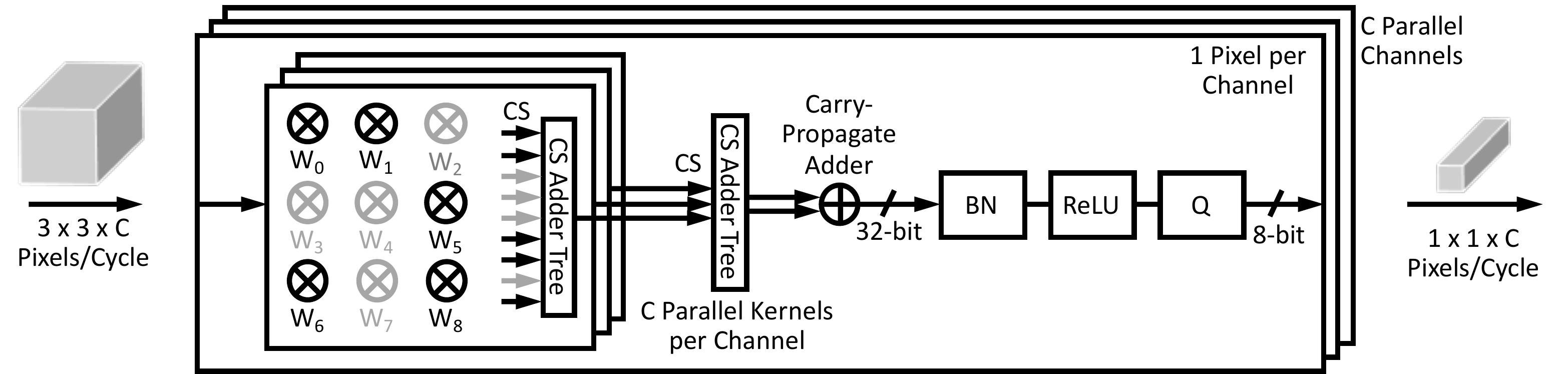}
\vspace{-16pt}
\caption{A fully-parallel fixed-weight native convolution hardware datapath stage for a $3 \times 3$ CONV layer.
Other CNN layer shapes are implemented in an identical fashion, but with different dimensions.
``CS'' denotes carry-save arithmetic representation.
``BN'' denotes batch normalization and incorporates the bias term.
``Q'' denotes a programmable quantization function that converts from 32-bit to 8-bit.
The multiplier symbols actually represent fixed-weight shift-add scalers with a single input operand.
Grey multipliers and signals denote hardware removed due to pruned zero or small non-zero weights.}
\label{fig:datapath_hw}
\vspace{6pt}
\end{figure*}

\textbf{Transfer Learning and Domain Adaptation.}
In FixyNN, we use transfer learning techniques to share an optimized fixed feature extractor amongst multiple different back-end CNN models.
\cite{yosinski_nips18} first established the transferability of features in a deep CNN, outlining that the early layers of a CNN learn generic features that can be transferred to a wide range of related tasks. 
Fine-tuning the model on the new task yields better performance~\cite{yosinski_nips18} than training from scratch. 
Transfer learning has subsequently found a wide range of applications. 
For example, a deep CNN trained on the \textbf{ImageNet} dataset~\cite{russakovsky2015imagenet} was successfully transferred to detect pavement distress in roads~\cite{gopalakrishnan2017deep}. 
Interestingly, more recent work demonstrated it is also possible to fix the last fully-connected layer in a CNN as a Hadamard matrix~\cite{Hoffer_18}.

Domain adaptation~\cite{tzeng2015simultaneous} is a concept closely related to transfer learning. 
It refers to learning adaptive models that work on different visual domains (e.g. hand-written digits versus printed street numbers). 
The residual adapter architecture~\cite{rebuffi_17,rebuffi2018efficient} marks the recent progress in this field to efficiently learn parametrized models for several tasks and domains simultaneously.
FixyNN can benefit from future advances in transfer learning and domain adaptation techniques.


\textbf{Hardware Generators for CNN Accelerators.}
A number of previous works have proposed solutions to 
automatically generate optimized hardware accelerator designs~\cite{venieris_18,tabla,sharma2016micro,lobato_nips16,reagen_islped17}.
There are also some relevant contributions from the image processing
domain~\cite{halide,darkroom} that similarly generate high-performance convolution hardware.
The DeepFreeze tool we developed in this work was a necessity in order to explore
fixed-weight feature extractors, as hand-written Verilog modules containing millions of parameters
would have been impractical otherwise.
We did not explore applying FixyNN on FPGAs~\cite{finn} in this paper, but plan to look at this in future work.
We are also planning to explore heavily-constrained Internet-of-Things (IoT) applications~\cite{kodali_iccd17} in future work.



\section{Fixed-Weight Feature Extractor Hardware Design}
\label{sec:hw}

\begin{figure*}[t]
\centering
\includegraphics[width=0.8\textwidth]{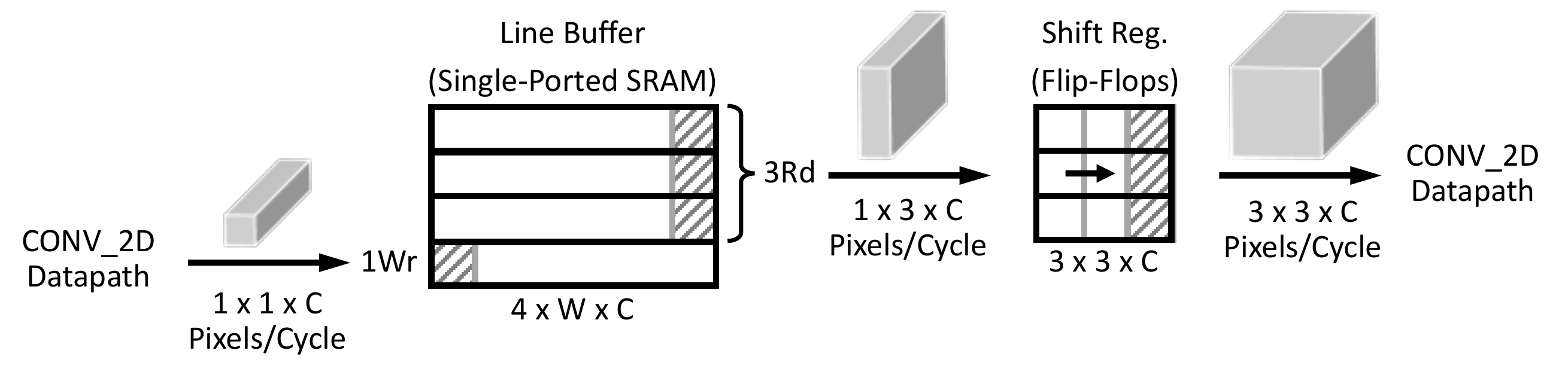}
\vspace{-10pt}
\caption{Overview of the fully-pipelined feature map buffering micro-architecture between consecutive layers of fixed-weight fully-parallel CNN layers.
This example illustrates the case for two consecutive CNN layers with 3$\times$3 kernels.}
\label{fig:buffering_hw}
\end{figure*}


FixyNN combines two specialized hardware accelerators: a heavily-optimized \textit{fixed-weight} feature extractor (FFE), and a more conventional \textit{programmable} CNN accelerator.
This combination provides very high energy efficiency without sacrificing generalization across a range of datasets.
Fixing the weights of a convolution (CONV) layer in a fully-parallel, fully-pipelined FFE accelerator enables a number of aggressive hardware optimizations in the FFE, and therefore results in significantly improved throughput and energy efficiency, which cannot be matched by a programmable accelerator.
We emphasize five major optimizations stemming from fixing weights in the hardware.
\begin{itemize}[itemsep=-1pt,topsep=-1pt,leftmargin=*]
\item \textbf{Fixed Shift-Add Scalers.} 
Hardware weight multipliers, which ordinarily have two input operands, are transformed into simple fixed scalers with a single input operand.
Fixed scalers are formed by simply adding a series of hard-coded bit-wise shifts of the input operands and are very cheap in hardware.
The number of bit-shifts and additions required per fixed multiplier is determined by the number of non-zero bits in the binary representation of the weight (i.e. Hamming weight).
This represents a very significant \textit{strength reduction} and results in substantial reduction in power consumption, logic delay and silicon area~\cite{cooper_2001_strengthreduction}.
\item \textbf{Zero-Overhead Weight Pruning.}
Weights with a zero or small non-zero value are redundant and can be explicitly removed from the datapath hardware.
This results in a reduction in datapath area and power, linearly proportional to the weight sparsity for the layer.
In a programmable CNN accelerator, there is overhead in exploiting sparsity, due to the requirement to encode the position in the matrix of non-zero weights~\cite{scnn}.
\item \textbf{Optimized Intermediate Precision}
The precision used for multipliers and accumulators are typically set to the worst-case values in a programmable accelerator.
However, in the FFE, we know the weights and their magnitude a-priori, and can therefore perform static analysis to optimize the product and accumulator bit-widths, which further reduces the hardware cost.
\item \textbf{Zero DRAM Bandwidth.}
The weights for the CONV layers implemented in the FFE are hard-coded in the datapath logic and do not need to be stored in memory.
Hence, unlike a programmable accelerator, there is no need to access expensive off-chip DRAM when using the FFE.
\item \textbf{Minimal Activation Storage.}
By using native convolution that does not incur storage overheads for \textit{IM2COL} expansion~\cite{warden_gemm}, and also implementing fully-pipelined hardware, we can reduce storage of activation feature maps to a minimum.
This is in contrast to programmable accelerators, which typically process layers in a serial fashion, to maximize weight reuse, and therefore must buffer the entire output feature map for each layer at once.
\end{itemize}

In the remainder of this section, we describe the hardware design of the FFE.
We first describe the arithmeric datapath stage, followed by the buffering stage, and finally the tool flow to automatically implement and optimize the FFE from a high-level model description.

\subsection{Fully-Parallel Fixed-Weight CNN Datapath}
\label{sec:hw:datapath}

The computation for each CONV layer is implemented as a flat, fully-parallel, pruned fixed-weight arithmetic logic stage (Figure~\ref{fig:datapath_hw}).
The fixed scalars that replace the multipliers are generated by the synthesis tool, as the weights 
are embedded as literals in the Verilog hardware description language (HDL).
These fixed scalars are also subsequently optimized by the synthesis tool to reduce gate-count, using techniques such as Booth recoding~\cite{booth_51}, canonical signed-digit encoding and other well-known datapath optimizations~\cite{zimmermann_2009}.
The adder trees following the multipliers are combined by the synthesis tool into a wide carry-save (CS) addition tree with a single carry-propagate adder~\cite{zimmermann_2009}.
Following the convolutions, there are operations in each layer for batch normalization (BN)~\footnote{A widely-adopted technique to improve performance and stability by ensuring layer outputs have zero mean and unit variance~\cite{ioffe_2015_batchnorm}.}, which scale and shift activations (and integrates the bias term), rectified linear unit (ReLU) activation function and a quantization step to convert from the wider precision of the accumulator node back to the narrow representation for activation data.
As we will describe in Section~\ref{sec:results:ml}, the BN parameters are important for transfer learning, so we keep these programmable, using dedicated registers.
This is not a big overhead as there are a very small number of BN parameters.
Simple max pooling layers are also supported.

\subsection{Fully-Pipelined CNN Buffering}
\label{sec:hw:buffering}

In contrast to programmable CNN accelerators that typically convert convolution into Generic Matrix Multiplication (GEMM), computing the CNN in a serial fashion, the FFE implements native convolution with \textit{fully-pipelined} CONV layers.
However, buffering is required between consecutive datapath stages, because a typical $3 \times 3 \times C$ CONV kernel, where $C$ is the number of channels, consumes a $3 \times 3 \times C$ input pixel tensor per cycle, but generates only a single small $1 \times 1 \times C$ output tensor, where $C$ is the number of output channels.
Hence, we must buffer several $1 \times 1 \times C$ outputs into a larger $3 \times 3 \times C$ input for the next layer.

This buffering function is achieved using the common approach of a \textit{line buffer}, which stores activations of each layer row by row until the required tensor size has been built up.
Figure~\ref{fig:buffering_hw} gives an overview of the arrangement for a simple CNN layer with a $3 \times 3$ kernel shape.
In this case, due to the discrepancy in input/output tensor dimensions, we need to buffer three full rows before we can start to generate the larger tensors we need for the following layer.
We implement the line buffer using simple single-port SRAMs, and therefore actually require four independent SRAM banks, such that we can write a single-row patch to one bank per cycle, and read the three-row patch from three banks per cycle, concurrently.
After reading/writing the last pixel in a row, the four banks are rotated to overwrite the data associated with the oldest row (double-buffer).
This arrangement can be further optimized~\cite{darkroom,rigel,halide}, for example, by using dual-port SRAMs, which were not available to us in our process technology.

Following the SRAM line buffer, a flip-flop based shift-register is implemented such that the convolution window moves efficiently over the feature map, without re-reading data.
The shift-register consumes $1 \times 3 \times C$ pixels per cycle from the SRAM line buffer and outputs a $3 \times 3 \times C$ pixel volume per cycle.
The advantage of the shift-register stage is an SRAM bandwidth reduction of $3 \times$.
Larger CNN kernels, such as $5 \times 5 \times C$ and $7 \times 7 \times C$ are arranged in a similar fashion, with dimensions scaled appropriately.
Strides of more than one are also supported.
We also make a provision to allow the activation data to be optionally streamed from any intermediate buffer stage, to allow a smaller number of fixed layers to be utilized for models that are more difficult to train via transfer learning.


\subsection{DeepFreeze Tool Flow}
\label{sec:hw:DeepFreeze}

\begin{figure}[t]
\centering
\includegraphics[width=0.45\textwidth]{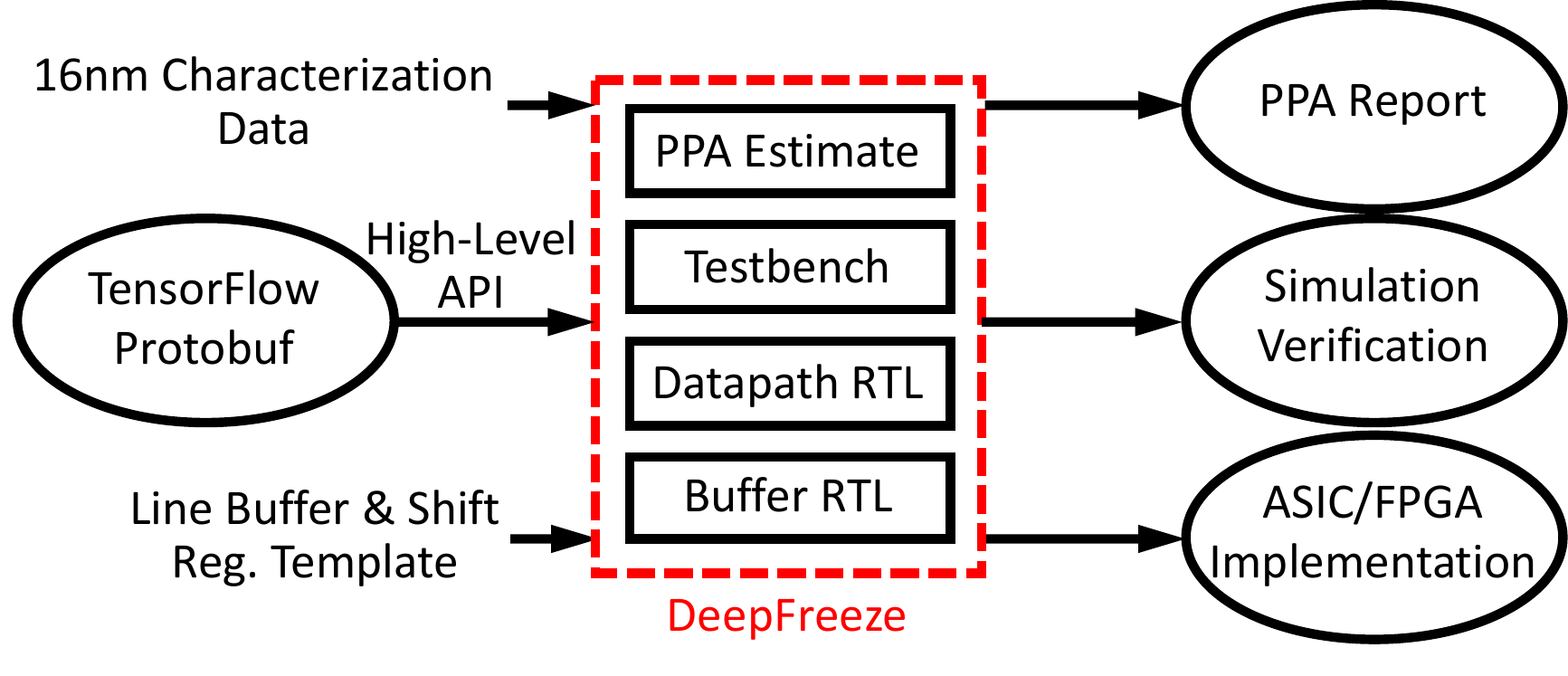}
\vspace{-10pt}
\caption{The DeepFreeze tool flow automatically generates Verilog HDL for optimized fixed feature extractors from a high-level description of the model in a software framework such as TensorFlow.}
\label{fig:DeepFreeze}
\vspace{-10pt}
\end{figure}

To facilitate implementing FFE accelerators with possibly millions of hard-coded weights, we developed an open-source tool called \textit{DeepFreeze}.
DeepFreeze generates fixed CNN hardware accelerator designs for a specified set of layers from a model described in a standard machine learning software framework, such as TensorFlow.

DeepFreeze first parses the network from a given framework into an internal representation of that model. It then generates a fixed datapath from the model description using a direct code generation step, which reads the model weights and emits Verilog source code with the weights embedded as immediate values.
Zero weights are automatically removed entirely from the hardware (pruning is assumed to be performed outside of the DeepFreeze tool-flow).
During the datapath generation, the bit-widths of the fixed scalars are optimized individually based on the scalar value.
The precision for the intermediate activations is specified as a hardware parameter, along with the accumulator width.
The final Verilog is constructed by connecting consecutive combinational datapath stages with buffer stages, which are instantiated from a parameterized Verilog template.
The generated Verilog can be directly read in by any synthesis tool for ASIC or FPGA implementation.
DeepFreeze also generates a validation suite with testbench for simulation.
Finally, the tool generates an estimate of power, performance and area (PPA) for the high-level model provided.
This estimate uses simple extrapolations from data derived from implementation experiments, and is useful for rapid design space exploration.






\section{Transfer Learning with a Fixed Feature Extractor}
\label{sec:ml}

In the previous section, we described the hardware design of a fixed feature extractor accelerator that offers substantially better throughput/latency and energy compared to programmable CNN accelerators.
However, we do not propose to fix the whole network for two reasons.
Firstly, for large models, the silicon area of the fixed hardware accelerator would be prohibitive in most applications.
Secondly, fixing the whole network would make it impossible to change the task or dataset; it would essentially result in a single-function hardware accelerator.
Therefore, in FixyNN we propose to fix only a portion of the front-end of the network, and use a canonical programmable accelerator to process the remainder (Figure~\ref{fig:concept}).
The fixed portion provides a set of more universal CNN features specific to the application domain of vision tasks, whereas the programmable portion of the network specific to a given a dataset.
In this section, we briefly outline how to train arbitrary CNN vision models that incorporate a fixed feature extractor implemented a-priori.


\textit{Transfer learning} is a concept that we introduced in Section~\ref{sec:related}. Here, we highlight transfer learning as a concept that suggests it is perfectly feasible to train a new model that incorporates a fixed feature extractor, at least within the same application domain of CV.
As previously motivated, the central advantage is that the performance and power efficiency of the fixed feature extractor are significantly superior.
In addition, there are a number of auxiliary advantages, such as a significantly smaller model to store, maintain and update. 

The CNN model architecture we use in this work is MobileNetV1~\cite{MobileNetV1}, which is an efficient model designed for mobile computer vision. 
MobileNet exploits the efficient depth-wise separable convolution layer, which is composed of $M$ $3\times 3 \times 1$ depth-wise convolution filters ($M$ is the number of input channels) and $N$ $1\times 1 \times M$ point-wise convolution filters ($N$ is the number of output channels). 
A depth-wise separable convolution layer costs between $8 \times$ to $9 \times$ less computation than a traditional $3 \times 3$ kernel.
Additionally, MobileNet is a suitable architecture for FixyNN because the FFE can directly concatenate the depth-wise and point-wise kernels without any buffering, as the output dimensions of the depth-wise layer are the same as the input dimensions of the point-wise layer.
MobileNet has $13$ CONV layers in total, with a fully connected layer for final classification. 
The first CONV layer is a traditional convolution layer and the remaining 13 CONV layers are depth-wise separable layers. 
A width multiplication factor $\alpha$ ~\cite{MobileNetV1} is introduced to explore different size models with the same basic architecture. 
For a given layer in the baseline MobileNet that has $M$ input channels and $N$ output channels,
the same layer in MobileNet-$\alpha$ has $\alpha M$ input channels and $\alpha N$ output channels. 
The width multiplier value of $\alpha$ reduces the computational cost and parameters by roughly $\alpha^2$. 

The procedure for training an image classification model on a given dataset is as follows.
We start by assuming the fixed feature extractor has already been defined, using the MobileNet architecture trained on the \textbf{ImageNet} data.
The early-layer weights are fixed for the feature extractor, while the remainder of the network is fine-tuned on the target dataset.
Further details of the training procedure can be found in Section~\ref{sec:methodology:ml}.



As discussed in Section~\ref{sec:hw}, fixing the weights in the feature extractor leads to a number of optimizations that cannot be as easily exploited in a programmable accelerator.
We may gain further benefits in latency, energy and silicon area through more aggressive optimization of the CNN layers for the fixed feature extractor by forcing more sparsity and Hamming weight reduction during training and fine-tuning.

\section{Experimental Methodology}
\label{sec:methodology}

To evaluate FixyNN, we conduct experiments in both hardware modeling and transfer learning.
The hardware modeling experiments compare FixyNN against state-of-the-art hardware accelerator designs.
The transfer learning experiments evaluate generalization of a fixed feature extractor across a set of tasks.


\subsection{Hardware Modeling}
\label{sec:methodology:hw}

FixyNN consists of two hardware components: the FFE, and a programmable CNN accelerator.
The FFE is generated using our DeepFreeze tool (Section~\ref{sec:hw:DeepFreeze}).
We use 8-bit precision for weights and activation data, and 32-bit for accumulators.
For ASIC implementation experiments, we use Synopsys Design Compiler with TSMC 16nm FinFET process technology to characterize silicon area.
Timing analysis for throughput/latency is performed with Synposys PrimeTime.
All simulations use a clock frequency of 810 MHz.
Power characterization is performed using Synopsys PrimeTime PX with switching activity annotated from simulation trace data.

The programmable accelerator is based on published results for the NVIDIA Deep Learning Accelerator (NVDLA)~\cite{nvdla}.
NVDLA is a state-of-the-art open-source neural network accelerator, with Verilog RTL for hardware implementation and a TLM SystemC simulation model that can be used for software development, system integration, and testing.
NVDLA is configurable in terms of hardware resources.
Table~\ref{table:nvdla} summarizes the published performance of NVDLA in six nominal configurations.


\begin{table}[ht]
\centering
\resizebox{0.46\textwidth}{!}{
\begin{tabular}{c c c c c c}
\hline
Config. & \#MACs & Buffer (KB)	& 16nm Area (mm\textsuperscript{2})	& TOPS	& TOPS/W \\
\hline\hline
A	& 64		& 128		& 0.55		& 0.056		& 2.0 \\
B	& 128		& 256		& 0.84		& 0.156		& 3.8 \\
C	& 256		& 256		& 1.00		& 0.358		& 5.6 \\
D	& 512		& 256		& 1.40		& 0.728		& 6.8 \\
E	& 1024		& 256		& 1.80		& 1.166		& 6.3 \\
F	& 2048		& 512		& 3.30		& 2.095		& 5.4 \\
\hline
\end{tabular}}
\caption{Published NVDLA configurations, reproduced from~\cite{nvdla}.}
\label{table:nvdla}
\end{table}

To explore the final FixyNN design space (Section~\ref{sec:results:hw}), we combine PPA models of an FFE containing the first \textit{N} layers of the network, along with the NVDLA programmable accelerator drawn from the published configurations.
DeepFreeze is used to model the PPA of the fixed feature extractor.
Since the hardware performance of the FFE is heavily dependent on the sparsity of the network, we assume a cautious 50\% sparsity across the model for simplicity. 
Prior work has demonstrated that 50\% of weights can be pruned from MobileNet with minimal accuracy loss~\cite{prune_gupta17}.
The hardware modeling of NVDLA is from published data.
Because the latency of the FFE is much lower than that of the programmable NVDLA in the configurations we tested, we assume perfect clock gating in FixyNN to eliminate FFE power when idle.
Finally, we do not model FC layers as they are heavily memory bound and we would never be able to fix them anyway due to the huge number of parameters.


\subsection{Transfer Learning}
\label{sec:methodology:ml}


The fixed feature extractor is constrained not only by silicon area considerations, but also by the achievable model accuracy.
The foundational work on transfer learning showed that as more layers are transfered, the accuracy becomes limited due to change in representational power and the later layers are more task specific than the early layers~\cite{yosinski_nips18}. 
In previous work, transfer learning is typically applied on big models such as AlexNet, which is prohibitively expensive from a hardware implementation point of view. 
Furthermore, it is arguably easier to perform transfer learning when the model capacity is very high as more parameters are available to fit the new dataset. 
In this paper, we perform a set of transfer learning experiments showing good performance with fixed weights on MobileNet, a much more constrained model.

Inspired by the \textit{visual decathlon challenge}~\cite{rebuffi_17} introduced to explore multiple-domain learning for image recognition, we choose seven different image recognition tasks to design our experiments: \textbf{ImageNet}~\cite{russakovsky2015imagenet}, \textbf{CIFAR-100}~\cite{krizhevsky2009learning}, \textbf{CIFAR-10}~\cite{krizhevsky2009learning}, \textbf{Street View House Numbers (SVHN)}~\cite{netzer2011reading}, \textbf{Flowers102 (Flwr)}~\cite{nilsback2008automated}, \textbf{FGVC-Aircraft (Airc) Benchmark}~\cite{maji2013fine}, and \textbf{The German Traffic Sign Recognition (GTSR) Benchmark}~\cite{stallkamp2012man}. 
These datasets vary in number of images, resolution and granularity. For example, \textbf{ImageNet} and \textbf{CIFAR-100} are diverse datasets with a wide range of objects, while \textbf{Flwr} and \textbf{Airc} are fine-grained recognition tasks for specific vision domains of flowers and aircrafts respectively.

For the first set of experiments, we use MobileNet-$0.25$, an efficient model with only $41$ million MACs and $0.47$ million parameters. 
The model is first trained on \textbf{ImageNet} to an accuracy of $49.8\%$ (state-of-the-art for this small MobileNet model) and then transfered to the other six vision tasks. 
The baseline results are obtained by performing full-fledged fine-tuning, where all the parameters of the model are updated during fine-tuning on the new dataset. 
This is used as the baseline case for a model running on a programmable DLA. 
Six different FixyNN topologies are explored in these experiments, with different number of layers being fixed. 
In some topologies, all batch normalization layer scaling and bias parameters in the model are retrained on the new dataset. 
We call this configuration Adaptive Batch-Normalization (BN).

Stochastic gradient descent with an initial learning rate of $0.01$ and momentum of $0.9$ is used to perform fine-tuning (except for GTSR dataset, where an initial learning rate of $0.001$ is used for better convergence). 
The learning rate is decayed $10 \times$ every $100$ epochs (200 epochs for GTSR). 
A batch size of $128$ is used. The seven datasets come with different resolutions. 
For the purpose of standardization, all images are resized to $224\times224$ using bilinear interpolation. 
Data augmentation preproccessing is applied to all datasets. Random color distortion, flipping and cropping are applied. 
Horizontal left-right flipping is turned off for \textbf{SVHN} and \textbf{GTSR}, cropping ratio is also increased as these two datasets are street number and traffic sign photos. 
MobileNet-$0.25$ is a limited capacity model so little regularization is required. 
Weight decay of $4 \times 10^{-5}$ is used in fine-tuning ($4 \times 10^{-4}$ for GTSR).

To demonstrate generalization of this approach, a second set of experiments are carried out using MobileNet-$1.0$. 
MobileNet-$1.0$ has $569$ million MACs and $4.24$ million parameters, which is about $10\times$ bigger than MobileNet-$0.25$. 
It is trained on \textbf{ImageNet} to an accuracy of $70.9\%$. 
We only transfer this model to \textbf{CIFAR100} to showcase the similar trend of transfer learning performance for a bigger model. 

\section{Experimental Results}
\label{sec:results}


In this section, we first describe the hardware performance of FixyNN, then explore the CNN generalization performance and finally draw the two together with a discussion.

\subsection{Hardware}
\label{sec:results:hw}

\begin{figure}[t]
    \centering
    \subfloat[Throughput]{\centering\label{fig:fixed_v_prog:tops}\includegraphics[width=0.45\textwidth]{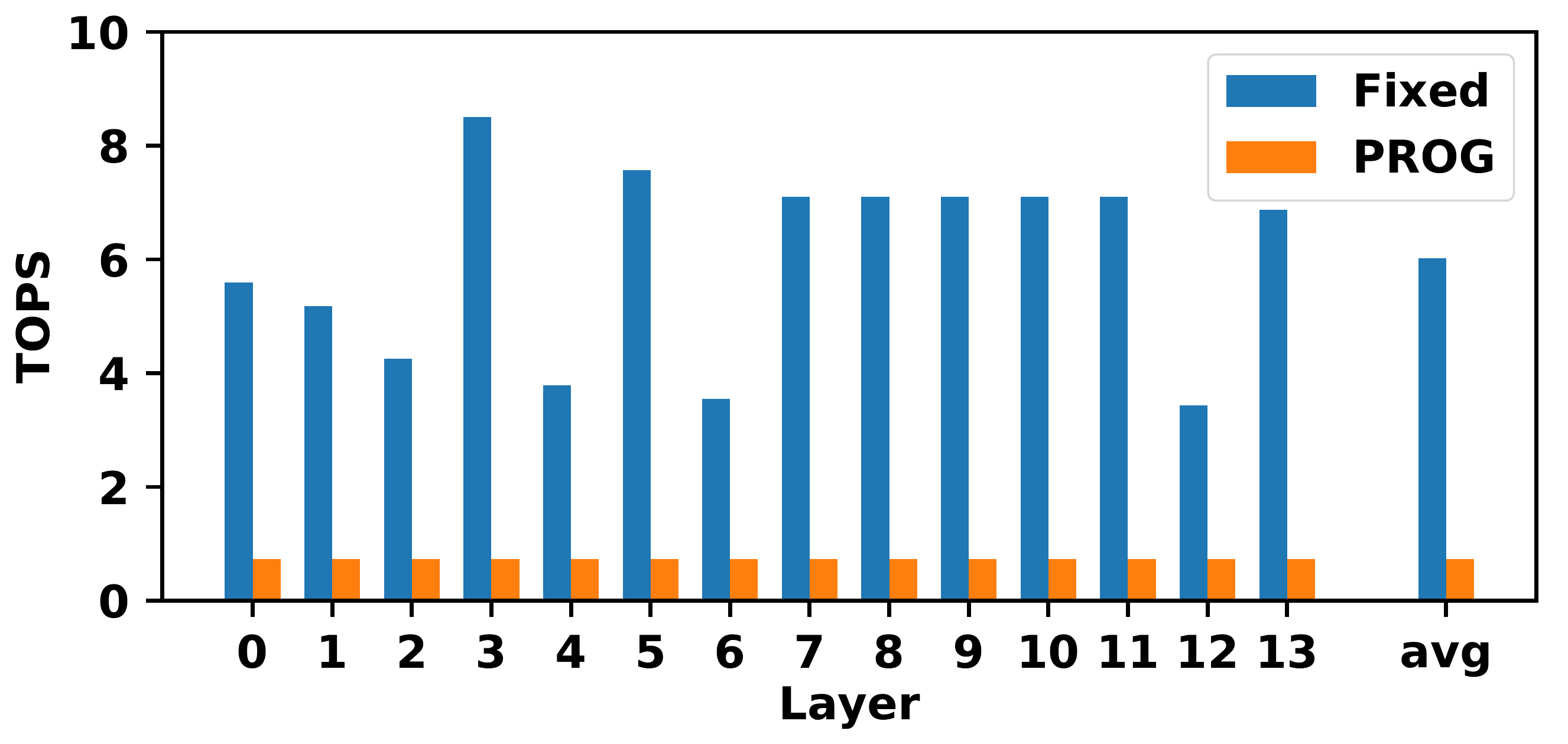}}

	\subfloat[Energy Efficiency]{\centering\label{fig:fixed_v_prog:topspw}\includegraphics[width=0.45\textwidth]{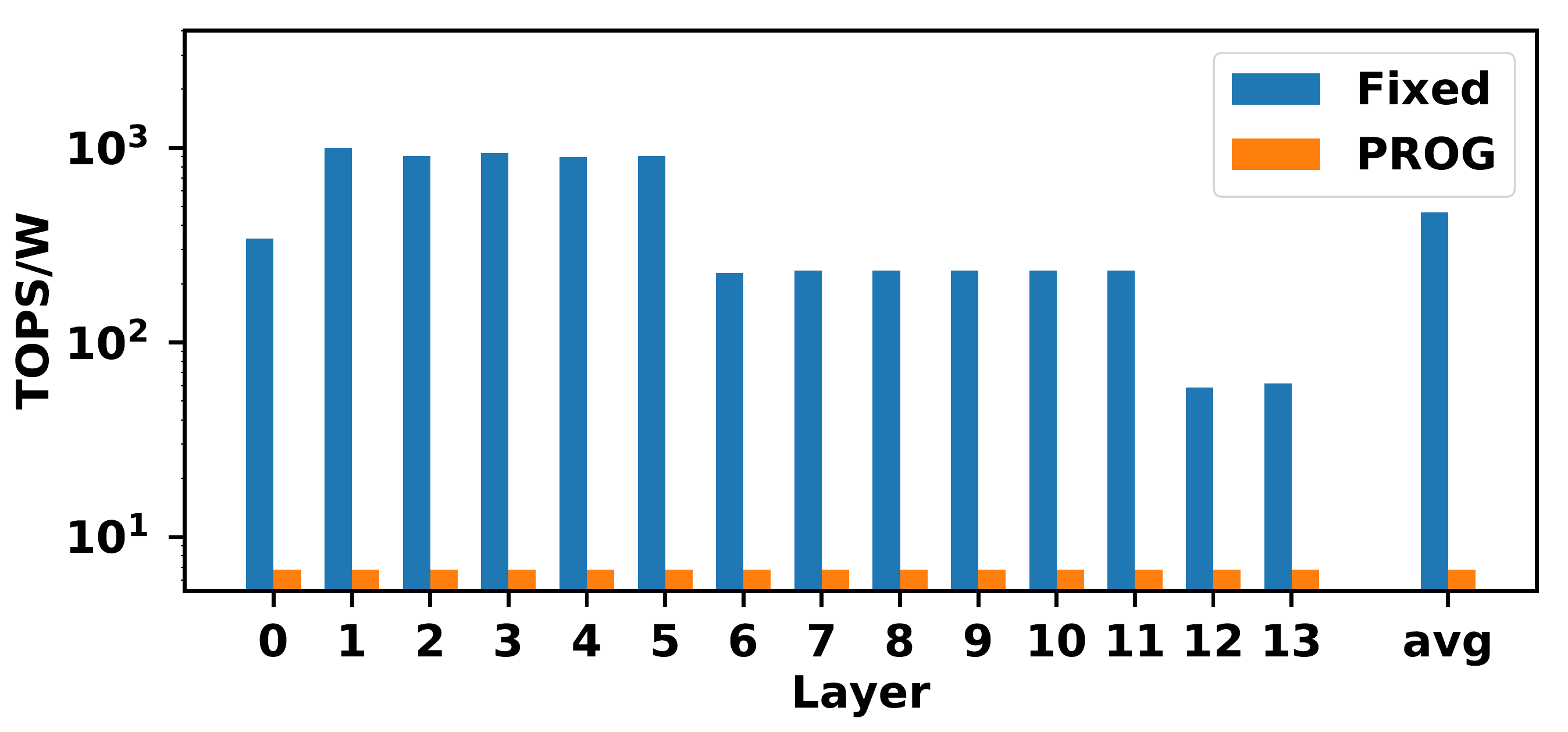}}
    \caption{Per-layer throughput and energy efficiency of a fixed-weight feature extractor vs programmable NVDLA on MobileNet-$0.25$.}
    \label{fig:fixed_v_prog}
\end{figure}

\begin{figure}[t]
	\centering
	\includegraphics[width=0.45\textwidth]{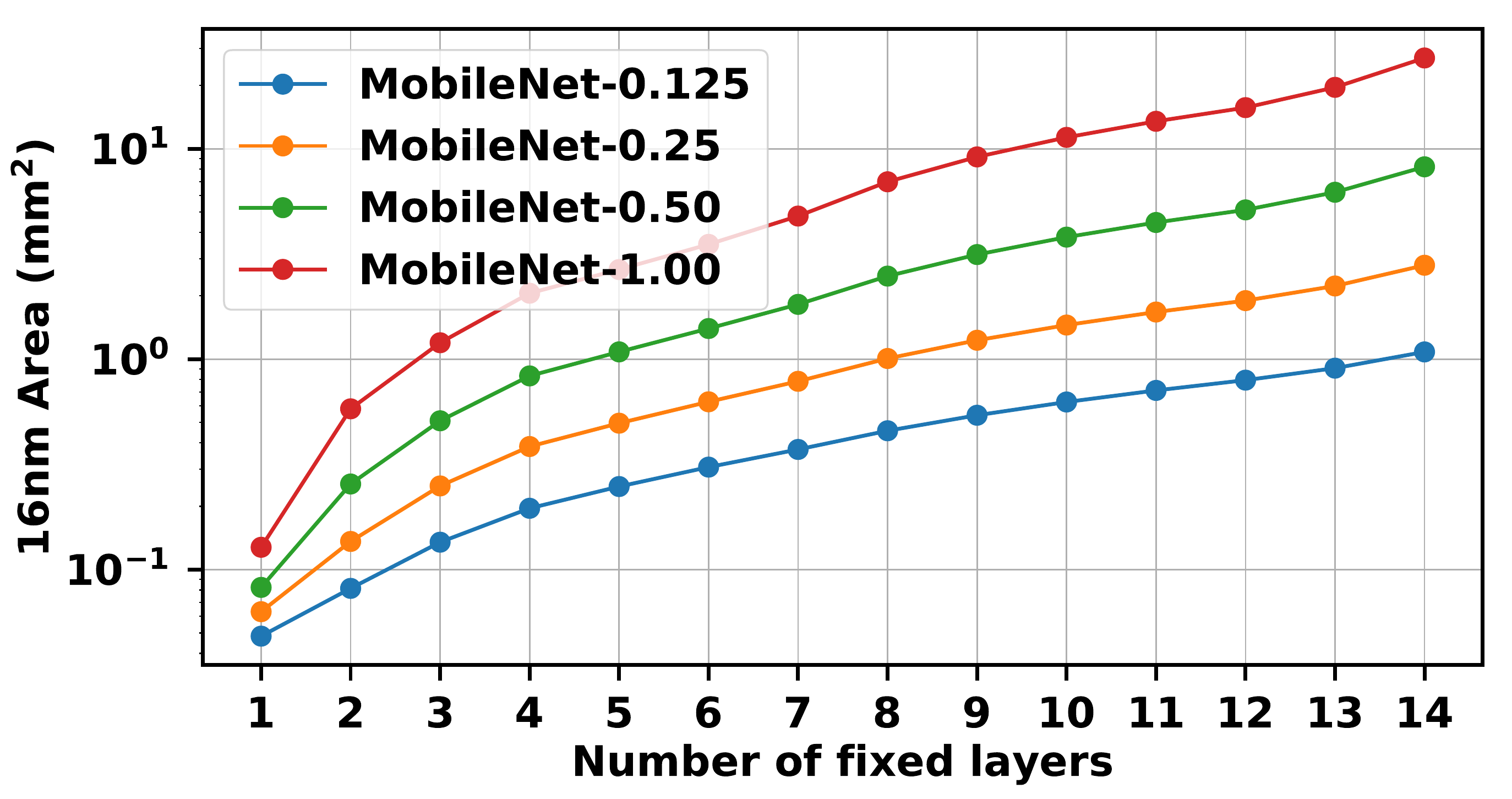}
	\caption{Cumulative area of a fixed feature extractor for MobileNets of varying width.}
	\label{fig:fixed_area}
\end{figure}

\begin{figure*}[t]
	\centering
  	\subfloat[Throughput]{\label{fig:hybrid:tops}\includegraphics[width=0.45\textwidth]{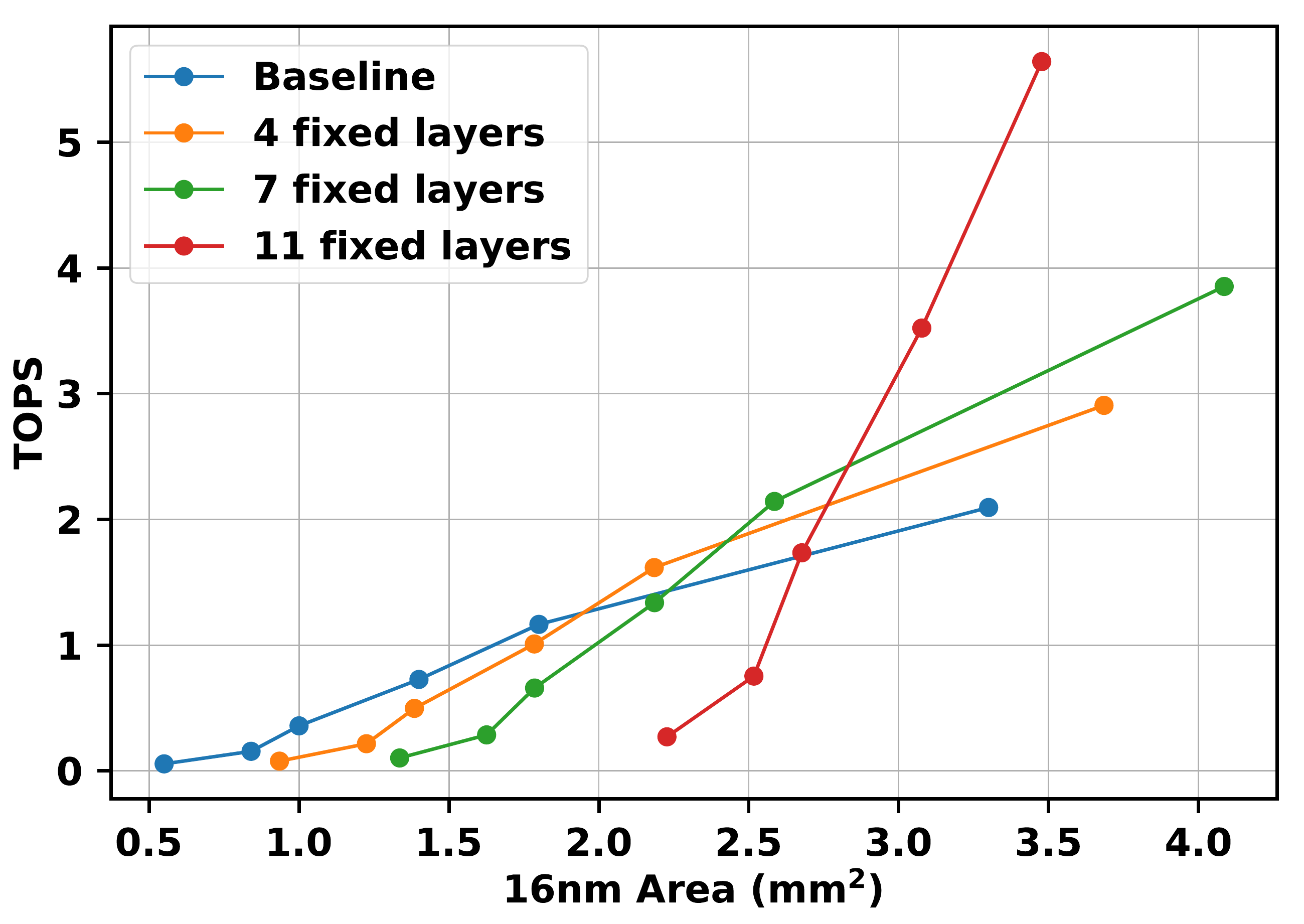}}
  	\subfloat[Energy Efficiency]{\label{fig:hybrid:topspw}\includegraphics[width=0.45\textwidth]{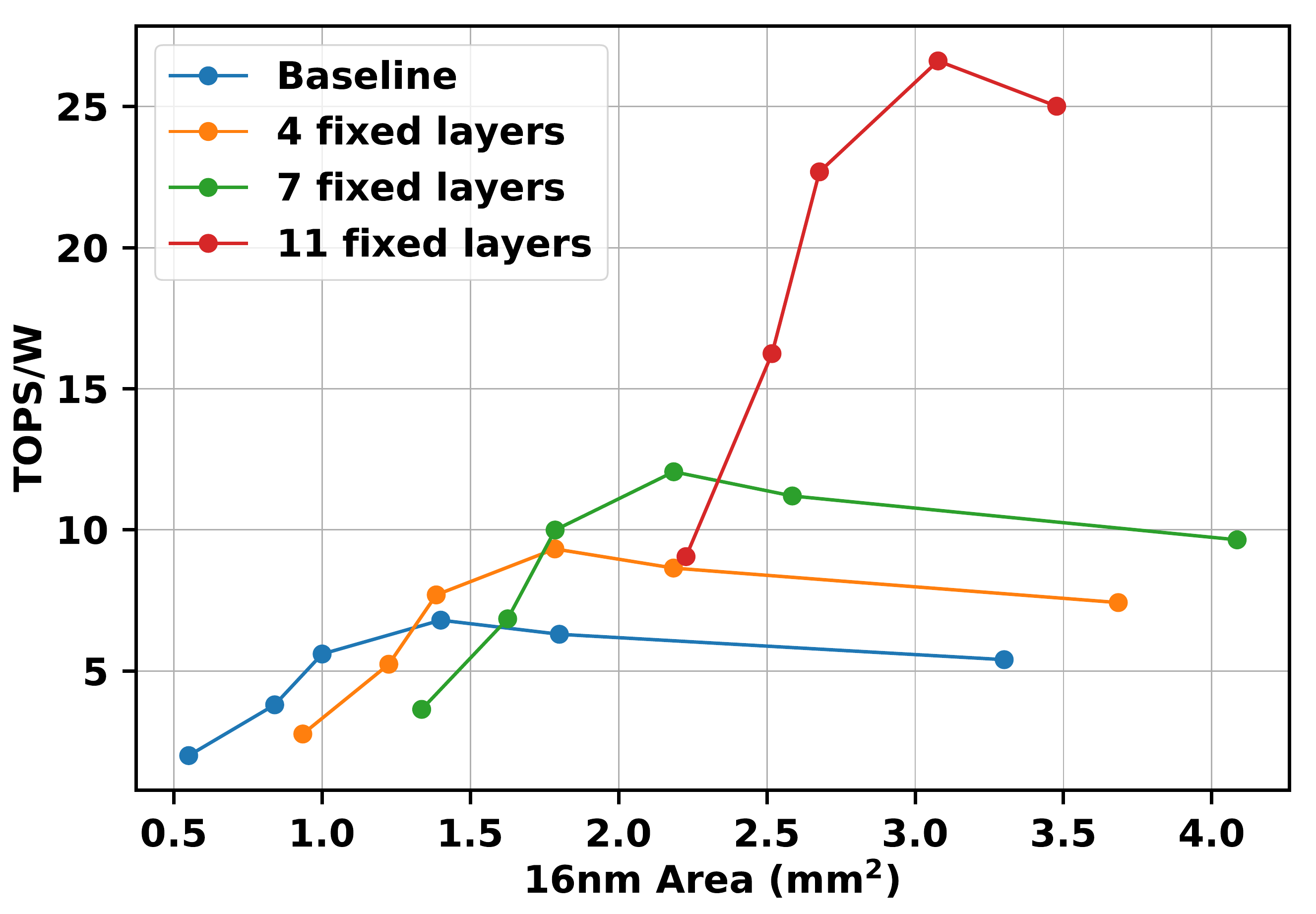}}
	\caption{Performance and energy efficiency of different FixyNN topologies. Each line corresponds to a single size feature extractor being used with different sized programmable accelerators. }
    \label{fig:hybrid}
\end{figure*}

\begin{table*}[t]
\centering
\resizebox{0.98\textwidth}{!}{
\begin{tabular}{c c | c c c c c | c c | c c}
\hline
\multicolumn{2}{c |}{Design Parameters} & \multicolumn{5}{| c |}{FixyNN} & \multicolumn{2}{| c |}{Baseline} & \multicolumn{2}{| c}{\textbf{Improvement}} \\
Priority & Area budget (mm$^2$) & Fixed layers	& NVDLA Config. & Total Area (mm$^2$) & TOPS	& TOPS/W & TOPS	& TOPS/W & \textbf{TOPS}	& \textbf{TOPS/W} \\
\hline\hline
\multirow{3}{*}{Throughput}
	& 2 		& None		& E 	& 1.80		& 1.17		& 6.30		& 1.17		& 6.30			& $\mathbf{1.00 \times}$		& $1.00 \times$	\\
	& 3 		& 7		& E		& 2.59		& 2.14		& 11.20		& 1.66		& 5.83		& $\mathbf{1.29 \times}$		& $1.92 \times$	\\
	& 4 		& 11		& E		& 3.48		& 5.64		& 25.01		& 2.21		& 5.29		& $\mathbf{2.55 \times}$		& $4.73 \times$	\\
\hline
\multirow{3}{*}{Efficiency}
	& 2 		& 7		& C	& 1.79		& 0.66		& 9.99		& 1.15		& 6.31		& $0.57 \times$		& $\mathbf{1.58 \times}$	\\
	& 3 		& 11		& C	& 2.68		& 1.73		& 22.69		& 1.71		& 5.77		& $1.01 \times$		& $\mathbf{3.93 \times}$	\\
	& 4 		& 11		& D & 3.08		& 3.52		& 26.62		& 1.96		& 5.53		& $1.80 \times$		& $\mathbf{4.81 \times}$	\\
\hline
\end{tabular}}
\caption{Pareto-optimal FixyNN configurations for a given area budget, with throughput and efficiency priority. 
``Improvement'' is relative to an NVDLA configuration of comparable silicon area.
All results shown are modeled in 16nm CMOS technology.
}
\label{table:paretos}
\end{table*}


To demonstrate the advantages of incorporating a FFE into a system, we begin by comparing the two hardware components of FixyNN.
Figure~\ref{fig:fixed_v_prog} compares the throughput (TOPS) and energy efficiency (TOPS/W) for the FFE and programmable NVDLA accelerators over each of the 13 layers of MobileNet-0.25.
Clearly, FFE outperforms NVDLA in all regards, showing an average improvement in TOPS and TOPS/W of $8.3\times$ and $68.5\times$, respectively.
This healthy improvement is essentially the motivation for exploring the fixed feature extractor.
However, the silicon area required by the FFE is a practical limitation on the number of layers we can reasonably fix in the FFE.
Figure~\ref{fig:fixed_area} demonstrates how the area of the FFE scales with the number of fixed layers for several different size MobileNet networks. 
In FixyNN, we want to balance the distribution of layers between the FFE and the programmable accelerators to maximize energy efficiency and generalization (Section~\ref{sec:results:ml}), given silicon area constraints.

Having demonstrated the advantages of the fixed feature extractor on single individual layers, we now demonstrate a practical FixyNN system.
We define a search space of potential FixyNN systems by combining a fixed feature extractor of a given size, and a programmable DLA of a given configuration (Table~\ref{table:nvdla}).
The design space is given in Figure~\ref{fig:hybrid} for throughput and energy efficiency.
Each line in these plots is a different number of fixed layers, while each marker on each line is a different configuration of the programmable accelerator (Table~\ref{table:nvdla}).
Our baseline for comparison is a fully programmable NVDLA accelerator with no fixed layers, which represents the current state-of-the-art.

In terms of throughput (Figure~\ref{fig:hybrid:tops}), all configurations scale approximately linearly with area.
At small area budgets, the fully programmable baseline outperforms FixyNN, because the FFE is heavily bottlenecked by the programmable NVDLA, resulting in little benefit from the extra area consumed by the FFE.
However at higher area budgets, FixyNN can afford to fix more layers, resulting in reduced load on the programmable DLA and large gains in throughput.
In terms of energy efficiency (Figure~\ref{fig:hybrid:topspw}), the baseline NVDLA scales well with area initially, due to an increase in data re-use and other amortizations, however it saturates (and even falls off) as limitations on utilization or memory bandwidth prohibit further gains.
Due to the exceptional energy efficiency of the FFE, as the load diverted from the NVDLA to the FFE increases, so too does the energy efficiency.
This becomes significant at area budgets greater than 1mm\textsuperscript{2}, at which point it becomes more efficient to utilize silicon area to fix more layers of the network than it is to scaling up the programmable accelerator.

An additional advantage of the FFE is the fact that it does not require access to expensive off-chip DRAM memory for either weights or activations, since weights are fixed in the datapath and activations are minimally pipelined in efficient and compact line buffers on-chip.
This saves power, and also sidesteps an important system-level constraint; NVDLA rapidly becomes bottlenecked on DRAM bandwidth as the accelerator is scaled up.


Table~\ref{table:paretos} gives pareto-optimal FixyNN configurations from the design space in Figure~\ref{fig:hybrid}, given different design constraints.
In general, this table shows it is more effective to implement a larger FFE at higher area budgets (above 1mm\textsuperscript{2}), as scaling the programmable NVDLA provides diminishing benefits beyond $\sim$1mm\textsuperscript{2}.
With an area budget of 4mm\textsuperscript{2}, FixyNN provides up to $2.55 \times$ and $5.84 \times$ improvement in TOPS and TOPS/W respectively, at iso-area for MobileNet-0.25.

\begin{figure}[t]
    \centering
    \subfloat[Area]{\label{fig:pie:area}\includegraphics[width=0.24\textwidth]{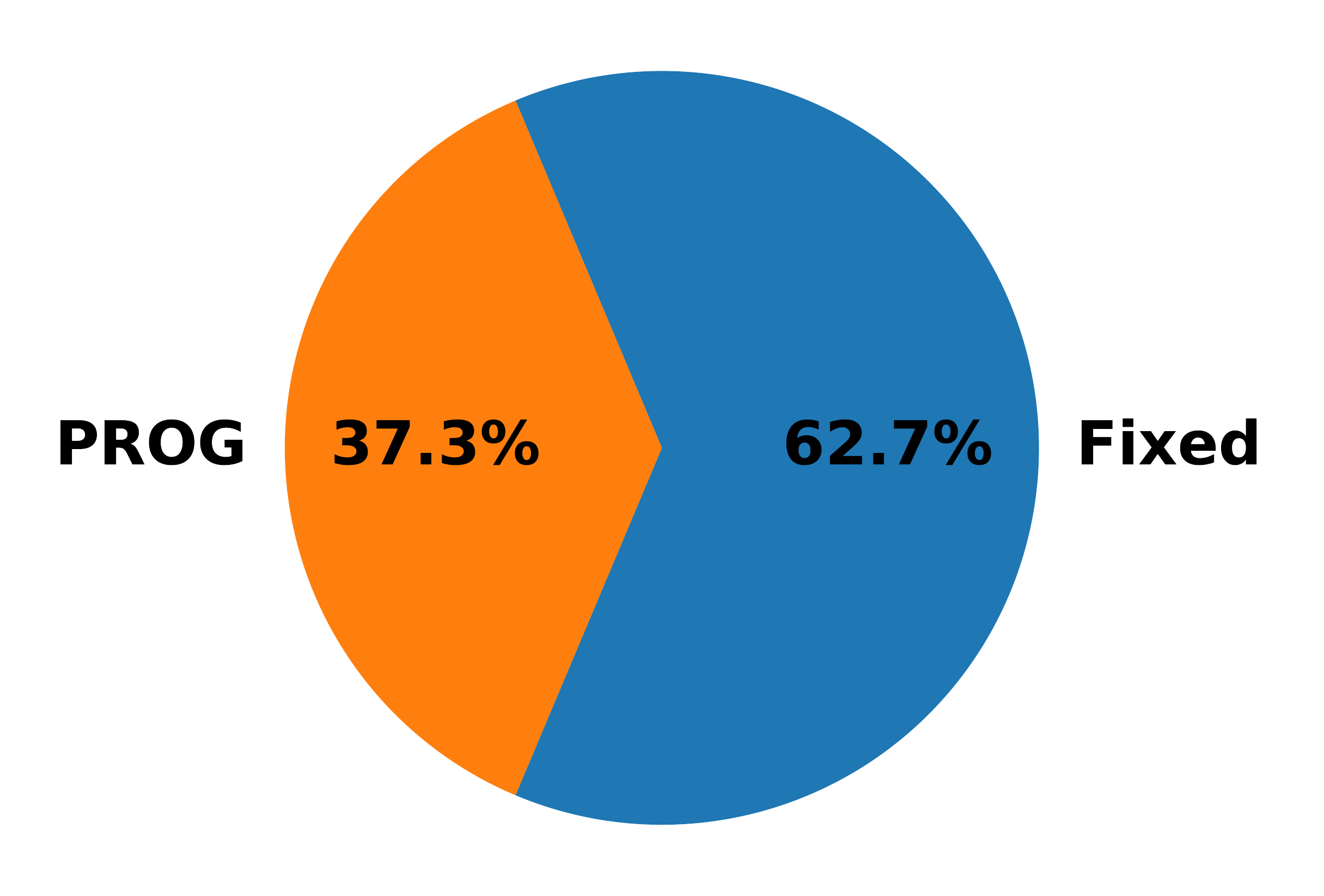}}
    \subfloat[Operations]{\label{fig:pie:ops}\includegraphics[width=0.24\textwidth]{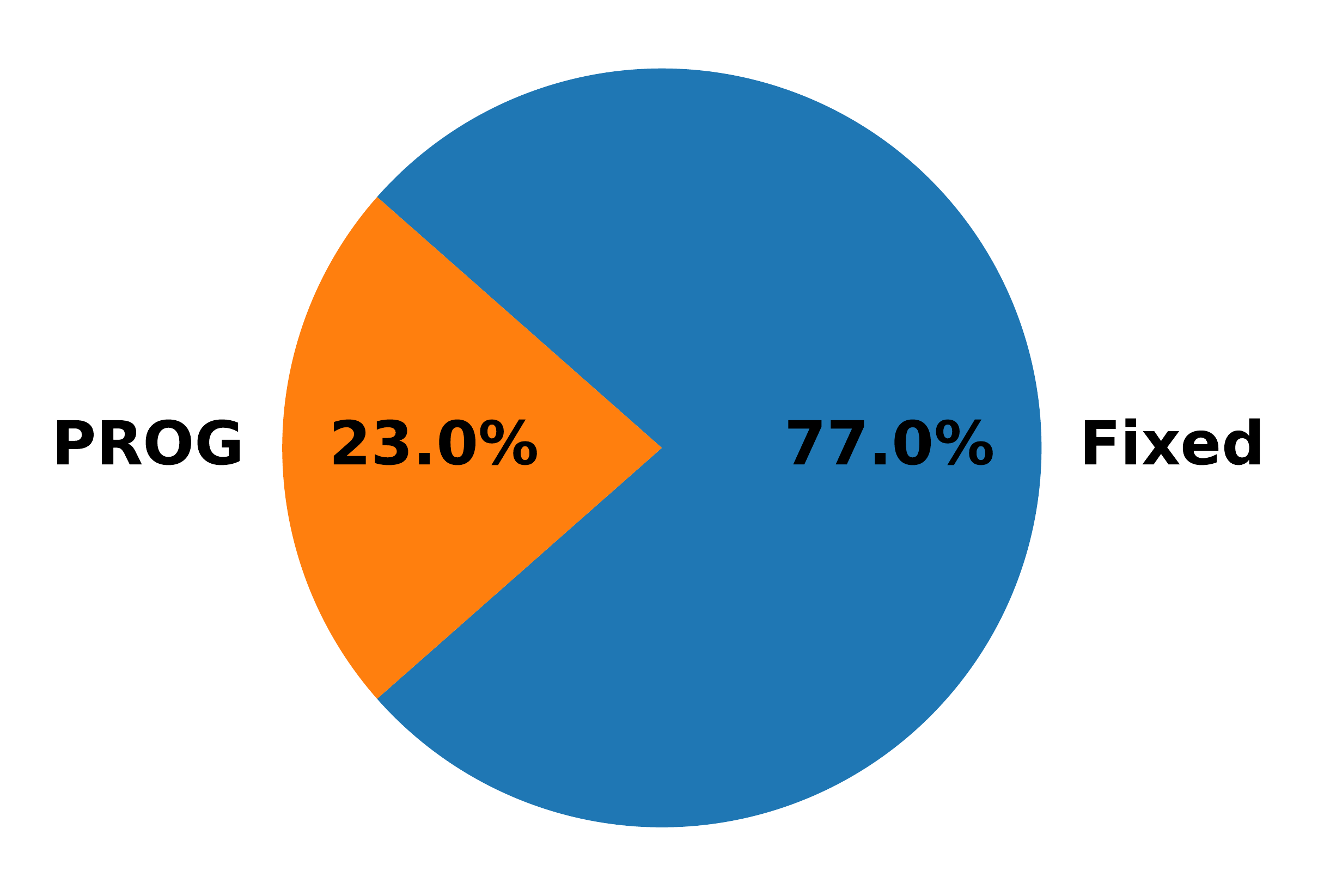}}

	\subfloat[Energy]{\label{fig:pie:energy}\includegraphics[width=0.24\textwidth]{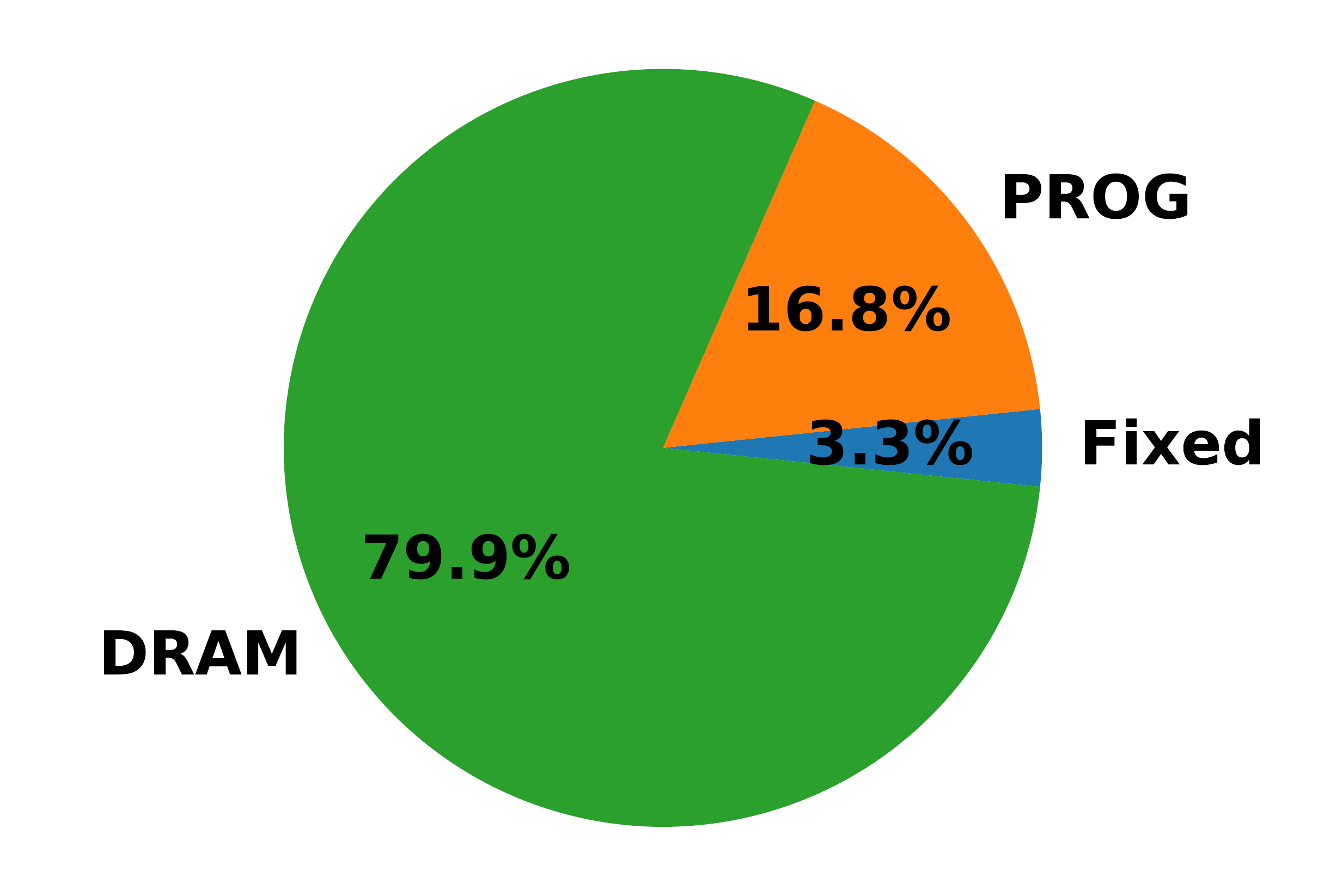}}
	\subfloat[Latency]{\label{fig:pie:latency}\includegraphics[width=0.24\textwidth]{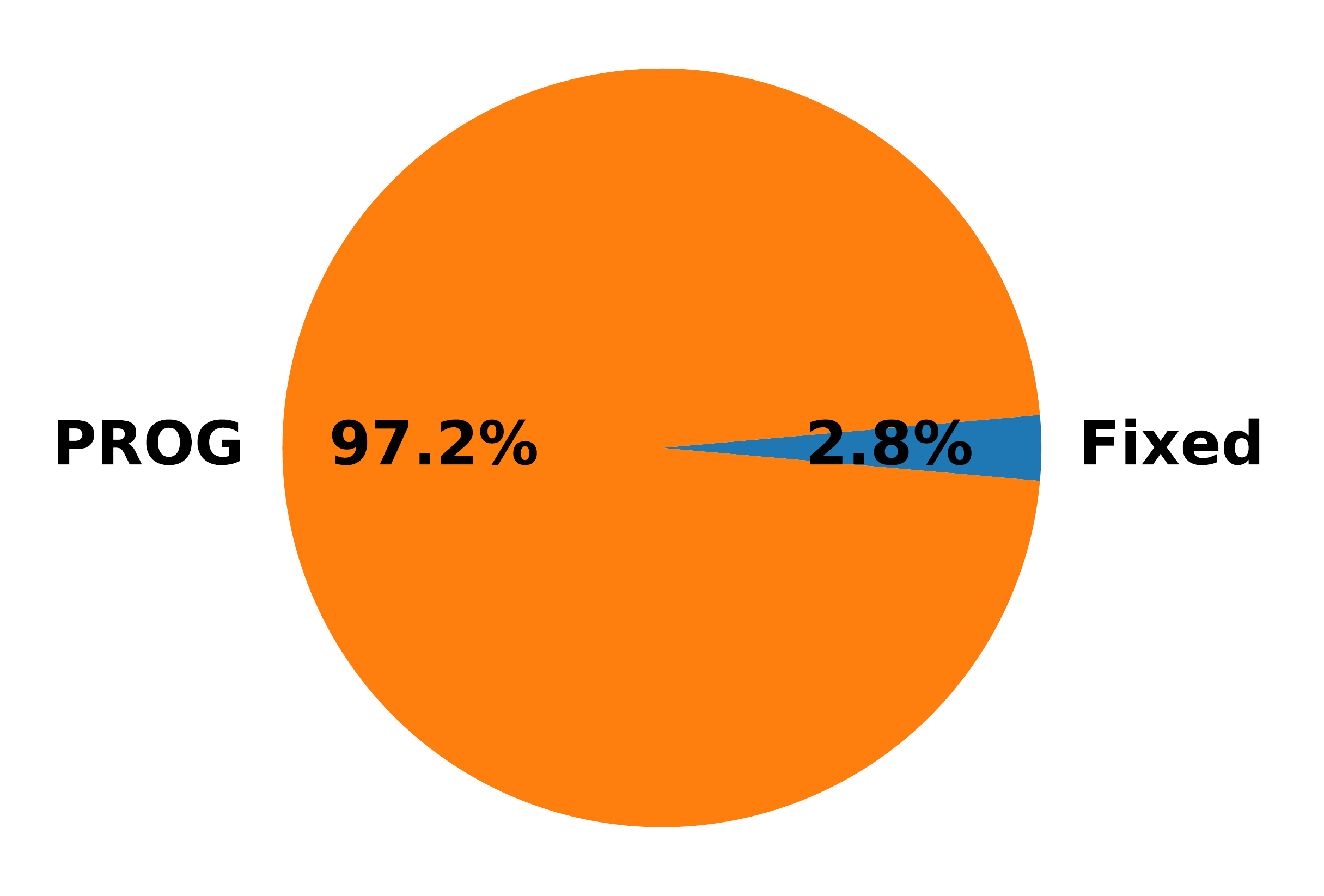}}
    \caption{PPA breakdown of FixyNN for MobileNet-$0.25$ with 7 fixed layers and a 1.00mm$^2$ NVDLA.}
    \label{fig:pie}
\end{figure}

We chose to investigate the optimal configuration for energy efficiency at an area budget of 2-3mm$^2$ (11 fixed layers with NVDLA configuration \textit{C}).
Figure~\ref{fig:pie} shows a breakdown of the PPA between the FFE and the programmable DLA.
This figure demonstrates how even though the fixed datapath performs a large majority of the operations in the network, it only takes a small fraction of the energy and latency that the programmable NVDLA requires.

The optimal configurations of FixyNN are dependent on the size of the model.
We repeated the experiment above, but using the larger MobileNet-1.00.
FixyNN now provides benfits at area budgets greater than 3mm\textsuperscript{2}, compared to the 1mm\textsuperscript{2} break-even point for MobileNet-0.25.
At an area budget of 4mm\textsuperscript{2}, fixing the first 4 layers of the network provides a $1.28 \times$ improvement in energy efficiency.
This improvement is even greater at larger areas.
The published results for NVDLA do not include any configuration larger than 3.3mm\textsuperscript{2}, and therefore it is difficult to make a fair evaluation at larger area budgets.
Nonetheless, we expect that as NVDLA scales up, memory bandwidth will bottleneck the system, resulting in reduced throughput and energy benefit.
FixyNN solves this problem by reducing the load on DRAM.

\subsection{Model Accuracy}
\label{sec:results:ml}

Table~\ref{table:ml_results_1} summarizes the accuracies for the first set of transfer learning experiments with MobileNet-$0.25$, where the first row shows the baseline accuracy. 
As we go down the table, a higher percentage of the network is fixed, hence a bigger FFE is used. Adaptive Batch Normalization helps a transferred model to achieve better accuracy with a relatively small hardware cost.
Images in different datasets come from different visual domains and have therefore very different statistical distributions, adaptive BN helps the model better adapt to the new domain.

Our experiments show that for datasets \textbf{CIFAR-100}, \textbf{CIFAR-10}, \textbf{SVHN} and \textbf{Flwr}, we can fix $77\%$ of the network while suffering less than $2\%$ loss in model accuracy. 
For datasets \textbf{Airc} and \textbf{GTSR}, similar accuracy performance relative to the baseline requires fixing a smaller percentage of the network in FFE (between $27\%$ and $44\%$). 

Transfer learning models are trained in floating-point datatype without forcing sparsity. Pruning and quantization are orthogonal to transfer learning and will affect model accuracy equally regardless of being transferred or not. Our observation for accuracy loss will hold even after further pruning and quantization of the model.

In Table~\ref{table:ml_results_2}, we report transfer learning accuracies for MobileNet-$1.0$. Only results on \textbf{CIFAR-100} are shown here. 
Similar trend in transfer learning accuracy loss is observed. 
Overall accuracies are improved as MobileNet-$1.0$ has a bigger model capacity. 
Fixing the first $11$ convolution layers of the network with adaptive BN results in $1.6\%$ accuracy drop. 

\begin{table*}[ht]
\resizebox{\textwidth}{!}{%
\begin{tabular}{ccc|ccccccc}
\hline
\multicolumn{3}{c}{Model} & \multicolumn{7}{|c}{Accuracy on datasets (\%)}   \\  Fixed layers & Adaptive BN & Fixed Ops (\%) & ImageNet  & CIFAR100   & CIFAR10    & SVHN   & Flwr   & Airc   & GTSR\\ \hline\hline
0 & N &0.0  & 49.8 & 72.8 & 93.5 & 95.8 & 88.1 & 67.7 & 97.7  \\ 
4 & Y & 27.1  & 49.8 & 72.5 & 93.3 & 95.7 & 88.3 & 66.7 & 97.8  \\
7 & Y & 44.3  & 49.8 & 72.0 & 92.7 & 95.8 & 87.5 & 64.0 & 95.0  \\
7 & N & 46.6  & 49.8 & 69.4 & 91.7 & 94.7 & 85.2 & 63.2 & 93.5  \\ 
11     & Y & 77.0  & 49.8 & 71.1 & 91.7 & 94.6 & 86.9 & 56.7 & 89.2  \\
14      &Y & 97.0 & 49.8 & 68.5 & 85.3 & 91.0   & 82.8 & 41.9 & 59.3  \\
14             & N & 100.0 & 49.8 & 54.5 & 77.0 & 48.0 & 77.8 & 30.5 & 46.1  \\   
\hline
\end{tabular}}
\caption{Transfer learning results for MobileNet-$0.25$ with fixed feature extractor, the model is trained on \textbf{ImageNet} and transferred to six different vision tasks.}
\label{table:ml_results_1}
\end{table*}

\begin{table}[h]
\centering
\resizebox{\columnwidth}{!}{%
\begin{tabular}{ccc|cc}
\hline
\multicolumn{3}{c|}{Model} & \multicolumn{2}{c}{Accuracy (\%)} \\ 
Fixed layers & Adaptive BN & Fixed Ops(\%) & ImageNet & CIFAR100	\\ \hline \hline
0 & N &0.0   & 70.9   & 81.7        \\
4 & Y&21.4  & 70.9   & 81.2         \\
7 & Y&39.9  & 70.9   & 80.7         \\
7 & N&40.6  & 70.9   & 80.2         \\
11 & Y&76.4  & 70.9   & 80.1        \\
14 & Y&99.1 & 70.9   & 76.7        \\
14 & N &100 & 70.9   & 61.6       \\      \hline
\end{tabular}}
\caption{Transfer learning results for MobileNet-$1.0$ with fixed feature extractor. The model is trained on \textbf{ImageNet} and transferred to \textbf{CIFAR-100}.}
\label{table:ml_results_2}
\end{table}


\subsection{Discussion}

Having presented the experimental results, we finally draw together some conclusions regarding the design of FixyNN systems.
Summarizing Section~\ref{sec:results:hw}, we found that the hardware throughput and energy-efficiency gains of FixyNN outpaces the baseline of an iso-area programmable NVDLA accelerator at the same silicon area cost when we fix 7 or more layers of Mobilenet-0.25.
The hardware throughput and energy efficiency of FixyNN reach as high as 5.64 TOPS ($2.55 \times$ better than the iso-area NVDLA baseline) and 26.62 TOPS/W ($4.81 \times$ better than the iso-area NVDLA baseline) respectively, at an area budget of $<4$mm\textsuperscript{2}.
On the other hand, Section~\ref{sec:results:ml} demonstrates experimentally that as we fix more layers in the FFE, the task of training a new network incorporating the FFE on a different dataset becomes more challenging, and will generally incur an accuracy loss which depends on the dataset.
Therefore, in practice, the system designer must balance the requirements of throughput/energy-efficiency and accuracy across a variety of datasets.
While this is obviously a nuanced trade-off, we offer a straightforward analysis to help emphasize the potential benefit of the FixyNN.

We consider an arbitrary constraint that the maximum tolerable degradation in accuracy is no greater than 2\% on the suite of six transfered datasets we examined in Section~\ref{sec:results:ml}.
We also specify a $<$3mm\textsuperscript{2} silicon area budget for accelerating CV workloads.
A FixyNN system that fixes 4 layers (27.1\% Ops) with adaptive BN, a 0.38mm\textsuperscript{2} FFE and NVDLA config. \textit{E}, can meet this specification, with a total area of 2.18mm\textsuperscript{2}.
Over all six datasets we studied, this FixyNN configuration achieves a maximum accuracy degradation of no more than 1.0\%, with the most challenging being \textbf{Airc}.
If we compare this design to a baseline consisting of a larger NVDLA of the same silicon area as the total FixyNN design (2.18mm\textsuperscript{2}, we achieve an improvement in throughput of $1.15\times$ and in energy efficiency of $1.42\times$.

As discussed in Section~\ref{sec:results:ml}, two of the six datasets are significantly less tolerant to a large number of fixed layers, which limits the improvement we demonstrate in the previous scenario.
Therefore, to prioritize \textit{average} performance across all datasets while otherwise still meeting the same constraints, we modify the FixyNN design so that the datasets with high accuracy degradation only use a portion of a larger FFE.
This allows us to define a FixyNN system that fixes 7 layers with adaptive BN (44.3\% Ops / 0.79mm\textsuperscript{2} FFE) and uses NVDLA config. \textit{E}, for a total area of 2.59mm\textsuperscript{2}.
With this configuration, four of the six datasets utilize the entire FFE as before, resulting in an improvement in throughput of $1.29\times$ (2.14 TOPS) and in energy-efficiency of $1.92\times$ (11.19 TOPS/W) over a baseline design of the same area.
The two datasets with high accuracy degradation may opt to use only 4 layers of the FFE, resulting in $0.98\times$ and $1.48\times$ in throughput and energy-efficiency, respectively.




\section{Conclusion}
\label{sec:conclusion}

Real-time computer vision workloads on mobile devices demand extremely high energy-efficiency for CNN computations, which can only be achieved with specialized hardware.
This paper evaluates FixyNN as a solution derived from closer integration of computer systems and machine learning.
FixyNN achieves an optimal balance of energy-efficiency from processing part of the network with heavily customized hardware for CNN feature extraction, and generalization to different CV tasks by means of a programmable portion that is trained using transfer learning.
Our experimental evaluation demonstrates that FixyNN hardware can achieve very high energy efficiency of up to 26.6 TOPS/W ($4.81 \times$ better than iso-area programmable accelerator).
We considered a suite of six image classification problems, and found we can train models using transfer learning with an accuracy loss of $<1\%$, and achieving up to 11.2 TOPS/W, which is nearly $2 \times$ more efficient than a conventional programmable CNN accelerator of the same area.



\newpage
\bibliography{nns}
\bibliographystyle{sysml2019}

\end{document}